\documentclass[runningheads,a4paper]{article}

\usepackage{amssymb}
\usepackage{graphicx}
\usepackage{centernot}
\usepackage{tabularx}
\newcolumntype{C}{>{\centering\arraybackslash}X} 

\usepackage[ruled,vlined,linesnumbered]{algorithm2e}
\usepackage{url}
\usepackage{graphicx}
\usepackage{amsmath}
\usepackage{subfigure}
\usepackage{xcolor}
\let\llncssubparagraph\subparagraph
\let\subparagraph\paragraph
\usepackage[compact]{titlesec}
\let\subparagraph\llncssubparagraph

\newtheorem{thm}{Theorem}[section]
\newtheorem{lemma}[thm]{Lemma}
\newtheorem{proposition}[thm]{Proposition}
\newtheorem{definition}{Definition}[section]
\newtheorem{corollary}[thm]{Corollary}

\newcommand{\dajian}[1]{{#1}}
\newcommand{\paul}[1]{{#1}}

\newcommand{\ie}{i.e., }
\newcommand{\eg}{e.g., }

\usepackage{hyperref}

\begin{document}


\title{Finding Risk-Averse Shortest Path with Time-dependent Stochastic Costs}


%
%
\author{
Dajian Li\footnote{\href{mailto:dajianl@andrew.cmu.edu}{dajianl@andrew.cmu.edu}} \and Paul Weng\footnote{\href{mailto:paweng@cmu.edu}{paweng@cmu.edu},  \url{http://weng.fr}} \and Orkun Karabasoglu\footnote{\href{mailto:karabasoglu@cmu.edu}{karabasoglu@cmu.edu}, \url{http://karabasoglu.com}}
}
%


%
%

\maketitle

\begin{abstract}
In this paper, we tackle the problem of risk-averse route planning in a transportation network with time-dependent and stochastic costs.
To solve this problem, we propose an adaptation of the A* algorithm that accommodates any risk measure or decision criterion that is monotonic with first-order stochastic dominance.
We also present a case study of our algorithm on the Manhattan, NYC, transportation network.

\end{abstract}

\section{Introduction}



\paul{
Shortest path problems have been extensively studied as they are canonical problems that appear in many domains, for instance transportation \cite{BastDellingGoldbergMuller-HannemannPajorSandersWagnerWerneck15,DellingSandersSchultesWagner09}, artificial intelligence \cite{RussellNorvig03} or circuit design \cite{PeyerRautenBachVygen09} to cite a few.
The standard version of this problem can easily be solved with classic shortest path algorithms such as the Dijkstra algorithm \cite{Dijkstra59} or the A* algorithm \cite{HartNilssonRaphael68}. 

In this paper, we focus more particularly on route planning in transportation networks.
While classically route planning operates with deterministic information (\eg expected travel duration), with the advent of intelligent transportation systems that provide real-time and historical traffic data, it becomes possible to design route planning approaches that take into account the stochastic and time-dependent nature of traffic condition. 
Indeed, as more and more cities open the access to historical traffic data, it is now possible to estimate a probability distribution over durations for each street at different times of the day. 
Such information can then serve as input to determine "shortest" paths that takes into account the variability of durations.
}

\paul{
More specifically, in this paper we focus on building a risk-averse route planning system for drivers in networks with stochastic and time-dependent costs. 
For a given origin and destination positions, it determines a shortest risk-averse path with respect to a pre-specified risk measure or decision criterion.
With this system, a driver could not only plan their trip in advance, but also avoid possible congestions. 
Consequently, this system could help reduce in particular travel time, traffic congestion and as a consequence exhaust emissions. 
}

\paul{
The contributions of this paper are twofold.
First, we propose an adaption of the A* algorithm, which extends and unifies previous algorithms \cite{ChenLamSumaleeLiTam14,ParmentierMeunier14} for computing a risk-averse shortest path in transportation networks where costs are stochastic and time-dependent.
Our approach can accommodate any risk measure or decision criterion that is monotonic with respect to first-order stochastic dominance.
Second, we demonstrate our proposition in the Manhattan, NYC transportation network with Conditional Value-at-Risk as a risk measure.

The paper is structured as follows.
The next section discusses the related work.
Section~\ref{sec:background} recalls the standard shortest path problem and the A* algorithm.
Section~\ref{sec:problem} defines the time-dependent stochastic-cost shortest path problem tackled in this paper.
Section~\ref{sec:solution} presents an adapted version of the A* algorithm to solve our problem.
Section~\ref{sec:casestudy} demonstrates our solution algorithm to route planning in Manhattan, NYC.
Finally, we conclude in Section~\ref{sec:conclusion}.
}

\section{Related Work}

Over the past decades, much effort has been devoted to the solution of the shortest path problem and its many variants.

Classic shortest-path algorithms such as the Bellman-Ford algorithm \cite{Bellman58,Ford56,Moore59}, the Dijkstra algorithm \cite{Dijkstra59} or the A* algorithm \cite{HartNilssonRaphael68} have been proposed before 1970s to solve the static version of the problem where edge costs are scalar and constant. 
However, in route planning, drivers usually value more travel times than distances, which has several implications.
Edge costs are generally non-stationary, that is they are a function of time (\eg driving the same street during peak hours or during normal hours lead to different durations).
They also tend to be random, depending on traffic conditions and other drivers. 
For these reasons, those classic algorithms for static shortest-path problems need to be adapted to this more general setting.

On the one hand, many studies have considered the non-stationary case, \ie time-dependent shortest path problem (TDSPP). 
Dreyfus \cite{Dreyfus69} extended the Dijkstra algorithm to TDSPP and Goldberg et al. \cite{GoldbergHarrelson05} solved TDSPP with a variant of the A* algorithm. 
TDSPP has been proven to be solvable in polynomial time under the First In First Out (FIFO) property \paul{(\ie which forbids an earlier arrival time while traversing an edge at a later time)} \cite{KaufmanSmith93} while it reveals to be an NP-hard problem without the FIFO property \cite{OrdaRom90}. 

\paul{
On the other hand, since Frank \cite{Frank69} studied stochastic-cost shortest path problem (SSPP), extensive work has been done on this problem (\eg \cite{SigalPritskerSolberg80,NieWu09,GavrielHanasusantoKuhn12,ParmentierMeunier14}). 
Bertsekas et al. \cite{BertsekasTsitsiklis91} considered an even more general class of stochastic shortest path problems (where node transitions are stochastic) and modeled them as a Markov Decision Problem \cite{Puterman94}. }
In SSPP, some researchers, such as Nie et al. \cite{NieWu09}, aimed at determining a shortest path \paul{guaranteeing a given probability of arriving on time}.
Recently, Gavriel et al. \cite{GavrielHanasusantoKuhn12} and Parmentier et al. \cite{ParmentierMeunier14} investigated risk-averse versions of SSPP by considering different risk measures, such as conditional Value-at-Risk (CVaR) \cite{EmbrechtsKluppelbergMikosch97}. 
However, neither of them considered the case where costs are time-dependent.

Besides, some work also tackles problems where edge costs are both non-stationary and random. 
\paul{Fu et al. \cite{FuRilett98} considered the problem of expected shortest paths in dynamic and stochastic traffic networks.}
Chen et al. \cite{ChenLamSumaleeLiTam14} studied time-dependent stochastic shortest path problems and proposed an adapted A* algorithm with first-order stochastic dominance in order to compute a reliable shortest path. 
However, the risk measure they use is Value-at-Risk (VaR) \cite{Jorion06}, which may not always be the most suitable measure. 
In this paper, we propose a practical algorithm for any risk measure that is monotonic with respect to first-order stochastic dominance (as in \cite{ParmentierMeunier14}) and test it with CVaR, which may be considered a better criterion than VaR as it has better properties \cite{ArtznerDelbaenEberHeath99} and  takes into account not only VaR but also the tail distribution.


\section{Background}\label{sec:background}


We first recall the definition of the classic shortest path problem.
Let $G = (V, E)$ be a directed graph (\eg corresponding to a transportation network) where $V$ is a set of 
nodes (\eg intersections and landmarks in a city) and $E \subset V^2$ is a set of 
directed edges (\eg lanes of streets).
\paul{The set of successors of a node $n$ is denoted $E^+(n)$, \ie $E^+(n) = \{n' \in V \,|\, (n, n') \in E\}$.}
A path $\pi$ of length $k$ in $G$ is a sequence of $k$ edges in $E$: $(n_1, n_2), (n_2, n_3), \ldots, (n_k, n_{k+1})$.
For convenience, we write $\pi = (n_1, n_2, \ldots, n_{k+1})$.
A subpath of a path is a consecutive subsequence of edges of that path.

The edges of graph $G$ are assumed to be valued by a cost function $c : E \to \mathbb R$ (\eg representing the distance or duration of travel in an edge).
We assume that costs are non-negative.
By extension, the cost of a (sub)path $\pi$, denoted $c(\pi)$, is defined as the sum of the costs of the edges in that (sub)path.
\paul{Let $\pi_{on}$ be a subpath from node $o$ to node $n$ and $\pi_{nd}$ a subpath from node $n$ to node $d$.
We denote $\pi_{on} \oplus \pi_{nd}$ the path obtained from the concatenation of the two subpaths.
Obviously, $c(\pi_{on} \oplus \pi_{nd}) = c(\pi_{on}) + c(\pi_{nd})$.}

Let $o \in V$ (resp. $d \in V$) be an origin (resp. destination) node.
The shortest path problem consists in searching for the path starting from node $o$ and ending in node $d$ that has the lowest cost.
Many efficient algorithms, such as the Ford-Bellman algorithm \cite{Ford56,Bellman58} or the Dijkstra algorithm \cite{Dijkstra59}, have been proposed to solve this problem.
In the case of transportation networks where the number of nodes may be large, those algorithms, even though polynomial in the size of graph $G$ may become impractical.
In that case, the A* algorithm 
may help to determine a shortest path faster.


The A* algorithm, proposed by Hart et al. \cite{HartNilssonRaphael68}, has been widely accepted as an efficient algorithm to solve the shortest path problem. 
As it is well-known, we only recall its principle and not its pseudo-code for space reasons.
In this algorithm, an extra heuristic information is assumed to be given: for any node $n$, an estimation $h(n)$ of the cost of the shortest path from node $n$ to destination node $d$ is available.
For instance, in transportation networks, where path distances are minimized, $h(n)$ can be defined as the Euclidean distance from node $n$ to destination node $d$.
The A* algorithm 
finds a path from origin node $o$ to destination node $d$ by exploring a tree of (sub)paths following a best-first-search strategy. 
In order to choose the best subpath to extend, the A* algorithm usually maintains a priority queue $\mathcal O$ \paul{of nodes representing subpaths ending in those nodes.
The priority $f(n)$ of a subpath $\pi_{on}$ ending in a node $n$} is defined as the sum of the cost cumulated so far and the heuristic estimation, \paul{\ie
}
$f(n) = g(n) + h(n)$ 
where $f(n)$ represents an estimation of the cost of \paul{of a path to node $d$ whose subpath is $\pi_{on}$, $g(n)$ is the cost of $\pi_{on}$} and $h(n)$ is the heuristic estimation of the cost of a subpath from node $n$ to node $d$.

Heuristic function $h(n)$ plays a significant role in the A* algorithm by influencing the number of (sub)paths A* algorithm will examine. 
Besides, whether the A* algorithm can eventually find the shortest path in the graph depends on the selection of the heuristic function $h(n)$. 
In order to guarantee the soundness of the A* algorithm, $h(n)$ should satisfy the following inequality: $\forall n\in V$,
\begin{align}\label{eq:admissible}
h(n) \leq \min_{\pi_{nd}} c(\pi_{nd})
\end{align}
where $\pi_{nd}$ represents a subpath from node $n$ to node $d$. 
This property means that the heuristic information provided by $h(n)$ is a lower bound to the best possible cost to reach node $d$ from node $n$.
For instance, the heuristic function defined as the Euclidean distance is admissible.
A heuristic function that satisfies inequality~(\ref{eq:admissible}) is called an {\em admissible} heuristic function.


\section{Problem Statement}\label{sec:problem}

\paul{We start with some notations. 
For any random variable $X$, we denote $P_X$ its probability densition function (pdf), $F_X$ its cumulative distribution (\ie $F_{X}(c) = \int_{-\infty}^c P_{X}(x) dx$) and $F^{-1}_X$ the (pseudo)inverse of $F_X$ (\ie $F_X^{-1}(\alpha) = \inf\{c \in \mathbb R \,|\, F_X(c) \ge \alpha \}$).
}

In a real transportation network, the duration for \paul{traversing an edge (\ie portion of a street)} is stochastic and dynamic.
Such a network can be represented as a directed graph $G = (V, E)$ as before, however, edge costs are now time-dependent real random variables.
For an edge $(n, n')$, random variable \paul{$C_t(n,n')$} denotes the random cost of traversing that edge at time $t$.
We assume random costs take non-negative values (representing durations) and S-FIFO\footnote{
The SFIFO property states that for any confidence level $\alpha$, leaving later cannot lead to an earlier arrival time: $t \le t' \implies t + F^-1_{C_t}(\alpha) \le t' + F^-1_{C_{t'}}(\alpha)$
where $t, t'$ are departure times,  $C_t, C_{t'}$ random costs of an edge and $\alpha \in [0, 1]$.
} (Stochastic FIFO) \cite{NieWu09}, which is a natural property in transportation networks, holds.



For a path $\pi=(n_1, n_2, \ldots, n_{k+1})$, its cost $C_t(\pi)$ for a departure time $t$ is also a random variable defined as the sum of the random costs of its edges.
\paul{
It can be written recursively as follows:
\begin{align}
C_t(\pi) = C_t(\pi') + C_{t + C_t(\pi')}(n_k, n_{k+1})
\end{align}
where $\pi'=(n_1, n_2, \ldots, n_{k})$. 
In a similar fashion, the pdf of $C_t(\pi)$ can be written:} 
\begin{align}
P_{C_t(\pi)}(c) = \int_{-\infty}^{+\infty} P_{C_t(\pi')}(x) P_{C_{t+x}(n_k, n_{k+1})}(c-x) dx
\end{align}


\paul{
The problem we tackle in this paper can then be formulated:
given a risk-averse criterion or risk measure $\rho : \mathcal X \to \mathbb R$ (with $\mathcal X$ the set of real random variables), we search for the $\rho$-minimum path $\pi^*$ for a departure time $t$, \ie
\begin{align}
\rho(C_t(\pi^*)) = \min_{\pi} \rho(C_t(\pi))
\end{align}
We call $\pi^*$ a risk-averse shortest path.
We assume that criterion $\rho$ satisfies a consistency property that relates $\rho$ to the first-order stochastic dominance, which is a partial order defined over probability distributions \cite{ShakedShanthikumar94}.
}

\begin{figure}[tb!]
\centering \includegraphics[width=0.5\linewidth]{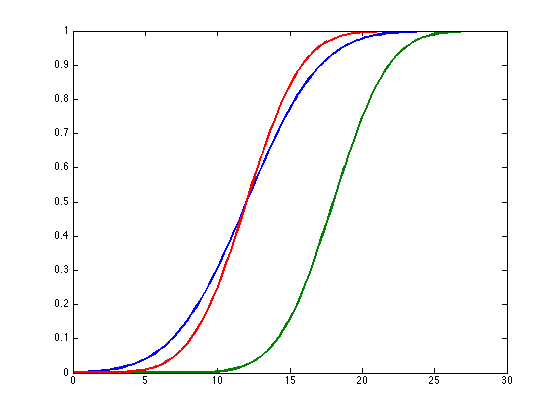}
\caption{Illustration of First-Order Stochastic Dominance: the green cumulative distributions FSD-dominates the blue and red ones, while the latter two are incomparable.}\label{fig:fsd}
\end{figure}

\paul{
\begin{definition}
{\em First-order stochastic dominance} (FSD) is defined as follows: 
Let $F_1$, $F_2$ be two cumulative distributions, $F_1$ (weakly) first-order-stochastically dominates (or FSD-dominates) $F_2$, denoted $F_1 \succsim_{FSD} F_2 $, iff $\forall x, F_1(x) \leq F_2(x)$. 
An illustration of FSD is shown in Figure~\ref{fig:fsd}. 
\end{definition}

The consistency property that we assume states that $\rho$ is monotonic with respect to first-order stochastic dominance:\\[.7ex]
{\bf FSD} $F_X \succsim_{FSD} F_Y$ $\Rightarrow$ $\rho(X) \ge \rho(Y)$\\[.7ex]
where $X$ and $Y$ are two real random variables and $F_X$ and $F_Y$ are their respective cumulative distributions.
This property is important because it will allow us to prune in the adapted A* algorithm.

Let us introduce another property that states that $\rho$ is increasing with the addition of a non-negative random variable:\\[.7ex]
{\bf INC} $\rho(X) \le \rho(X+C)$\\[.7ex]
where $X$ and $C$ are two real random variables and $C$ takes non-negative values.
In our setting, this is a natural property as random variables represent durations.

To prove that FSD implies INC,
\dajian{
we first introduce a lemma\footnote{For space reasons, we do not include the proofs.}:
\begin{lemma}\label{lm:inc}
\paul{Let $X$ be a real random variable and $C$ be a non-negative real random variable. Then,
$F_{X+C} \succsim_{FSD} F_{X}$.}
\end{lemma}
}

\paul{
Then, as a direct consequence of Lemma \ref{lm:inc}, we obtain:
\begin{proposition}
If $\rho$ satisfies FSD, $\rho$ also satisfies INC.
\end{proposition}
}
}

%

%
%
%
%
%
%
%
%
%

\paul{
As illustrations of $\rho$, we present three examples, Value-at-Risk, Conditional Value-at-Risk and Expected Utility, which all satisfy FSD (and therefore INC). \\[.9ex]}
\paul{
{\em Example 1.~ Value-at-Risk} (VaR) \cite{Jorion06} is a widely-used risk measure in finance.
For a fixed $\alpha \in [0, 1]$, it represents the threshold loss value, such that the probability the loss on an investment exceeds this value is $\alpha$. 
Formally, in our context, it is defined by:
$VaR_\alpha(X) = F_X^{-1}(\alpha) = \inf \{ x \in \mathbb R \,|\, F_X(x) \ge \alpha \}$
In other terms, $VaR_\alpha(X)$ is defined in our context as the $\alpha$-quantile of random variable $X$.
It is well-known that VaR satisfies FSD \cite{BauerleMuller05}. \\[.8ex]
}
\paul{
{\em Example 2.~ Conditional Value-at-Risk} (CVaR) \cite{EmbrechtsKluppelbergMikosch97}, also called {\em Expected Shortfall} is a risk measure that refines VaR. 
Because VaR is a threshold value (for a single fixed probability $\alpha$), it neglects the risk at the tail of the distribution.
CVaR remedies this shortcoming of VaR by measuring the expected loss at the tail above VaR. 
CVaR is mathematically defined by:
$CVaR_\alpha(X) = \mathbb E[X \,|\, X \geq VaR_\alpha(X)]$
where $X$ is a real random variable. 
The benefit of using CVaR instead of VaR is that CVaR takes into account not only the VaR value but also the tail information of a distribution. 
CVaR is known to satisfy FSD \cite{BauerleMuller05}. \\[.8ex]
}
\paul{
}
\paul{
{\em Example 3.~ Expected Utility} (EU) is a well-known decision criterion in decision under risk \cite{vonNeumannMorgenstern44} and decision under uncertainty \cite{Savage54}, which is known to satisfy FSD \cite{BauerleMuller05}.
It is defined as follows:
$EU(X) = \mathbb E(u(X))$
where $X$ is a real random variable and $u : \mathbb R \to \mathbb R$ is a so-called von Neumann-Morgenstern utility function. 
The utility value $u(x)$ represents how much $x$ is valuable. 
For this reason, function $u$ is assumed to be monotonic (\ie in our settings, $x \le y \Rightarrow u(x) \ge u(y)$).
In decision theory, it is well-known that a concave (resp. convex) function $u$ leads to a risk-averse (resp. risk-seeking) decision criterion.
Although we focus on risk-averse criteria in this paper (as it is most people's concern in transportation), note that our approach could also tackle the risk-seeking case.
}

\paul{
There are many other possible examples of $\rho$ that satisfies property FSD: for instance, semideviations \cite{OgryczakRuszczynski99}, rank-dependent utility \cite{Quiggin93}, Yaari's dual model \cite{Yaari87}...
Our solution algorithm covers all those cases.
}

\section{Solution Algorithm}\label{sec:solution}

\paul{We propose an algorithm that is an adapted version of the standard A* algorithm}
to solve the proposed risk-averse shortest path problem using time-dependent stochastic costs.
\paul{
It generalizes the algorithm proposed by Chen et al. \cite{ChenLamSumaleeLiTam14} to general $\rho$ measures that satisfies FSD and extends the algorithm proposed by Parmentier et al. \cite{ParmentierMeunier14} to the time-dependent cost setting. 
}

\begin{algorithm}[tb!]
\KwData{graph $G=(V, E)$, random costs $C_t$, heuristic $h$, criterion $\rho$, upperbound $UB$, origin node $o$, destination node $d$, departure time $t$}
\KwResult{risk-averse shortest path}
\Begin{
$\mathcal O \gets \{(o, 0)\}$ \\
\While{$\mathcal O \neq \emptyset$}{
   $(n, C) \gets $ highest priority pair in $\mathcal O$ \\
   \lIf{$n = d$}{return corresponding path}  \label{algol:end}
   remove $(n, C)$ from $\mathcal O$\\
   \For{$n' \in E^+(n)$}{ 
      $C' \gets C + C_{t+C}(n, n')$ \label{algol:sum} \qquad $\rhd${initial time $t + C$ selects the edge cost}\\
      $f(n', C') \gets \rho(C' + h(n'))$\\
      \eIf{$n' = d$ and $f(n', C') < UB$}{\label{algol:improveUB}
         $UB \gets f(n', C')$
      }{
         \lIf{$f(n', C') \ge UB$}{continue}
      }
      \eIf{$n' \notin \mathcal O$}{
         add $(n', C')$ in $\mathcal O$ 
      }{
         \eIf{$C'$ not FSD-dominating any $(n', C'') \in \mathcal O$ }{
            add $(n', C')$ in $\mathcal O$ and remove FSD-dominating $(n', C'') \in \mathcal O$
         }{
            continue
         }
      }
   }
}
return $\emptyset$
%
%
}
\caption{Proposed adapted A* algorithm}\label{algo:aastar}
\end{algorithm}

\paul{
The proposed algorithm keeps the basic features of the standard A* algorithm, for example, an open set $\mathcal O$ while adding new features such as labeling with random variables and path pruning using FSD dominance. 
}
\paul{
We now explain why the notions of label (for evaluating the value of a subpath ending in node $n$) and priority (for guiding the order the subpaths are examined) need to be redefined in our setting and how they can be redefined.
In the standard A* algorithm for computing a shortest path, a node $n$ in the priority queue $\mathcal O$ is the end node of a subpath for which only one label (\ie cumulated cost $g(n) = c(\pi_{on})$) needs to be stored.
This is possible because we have $\forall n \in V$:
\begin{align*}
&c(\pi_{on}) \le c(\pi'_{on}) \implies c(\pi_{on} \oplus \pi_{nd}) \le c(\pi'_{on} \oplus \pi_{nd})
\end{align*}
where $\pi_{on}$ and $\pi'_{on}$ are two paths from node $o$ to node $n$ and $\pi_{nd}$ is a path from $n$ to $d$.
Unfortunately, in our setting, a counterpart of these inequalities with respect to $\rho$ does not hold, due to the possible non-linearity of criterion $\rho$:
\begin{align}\label{eq:nmon}
\rho(C_t(\pi_{on})) \le \rho(C_t(\pi'_{on})) \centernot \implies \rho(C_t(\pi_{on} \oplus \pi_{nd})) \le \rho(C_t(\pi'_{on} \oplus \pi_{nd}))
\end{align}
In words, a dominated subpath can become non-dominated when extended. \\[.7ex]
{\em Example 4.~}
We give an example for the case when $\rho$ is VaR with $\alpha = 95\%$.
Assume the probability distributions are given in Table~\ref{tab:prob}.
One can check that:
\begin{align*}
&VaR(C_t(\pi_{on})) = 1 < 2 = VaR(C_t(\pi'_{on})) \mbox{ and } \\
&VaR(C_t(\pi_{on} \oplus \pi_{nd})) = 3 > 2= VaR(C_t(\pi'_{on} \oplus \pi_{nd}))
\end{align*}
\begin{table}[tb!]
  \begin{center}
    \caption{Cumulative distributions.}
    \label{tab:prob}
\begin{tabularx}{\linewidth}{X|XXXXX}
$x$ & 0 & 1 & 2 & 3 & 4\\
\hline
$F_{C_t(\pi_{on})}$ & 0 & 0.95 & 1 & 1 & 1\\
$F_{C_t(\pi'_{on})}$ & 0.9 & 0.9 & 1 & 1 & 1\\
$F_{C_t(\pi_{nd})}$ & 0.8 & 0.9 & 1 & 1 & 1\\
$F_{C_t(\pi_{on} \oplus \pi_{nd})}$ & 0 & 0.76 & 0.895 & 0.995 & 1\\
$F_{C_t(\pi'_{on} \oplus \pi_{nd})}$ & 0.72 & 0.81 & 0.98 & 0.99 & 1
\end{tabularx}
  \end{center}
\end{table}
}
\paul{
As a consequence of (\ref{eq:nmon}), labels have a more complex form.
Following previous related work \cite{NieWu09,ChenLamSumaleeLiTam14,ParmentierMeunier14}, the label of node $n$ is defined as $C_t(\pi_{on})$ instead of $\rho(C_t(\pi_{on}))$.
For a given node, two labels can be compared with FSD-dominance (thanks to Corollary~\ref{cor:fsd}).
As it is a partial order, a node can then receive several labels.
For this reason, elements of $\mathcal O$ are pairs $(n, C)$ where $n$ is a node and $C$ is a random variable representing the cost of a subpath from node $o$ to node $n$.

The priority of a pair $(n, C)$ in $\mathcal O$ is defined as $f(n, C) = \rho(C + h(n))$ where $h(n)$ is a known heuristic evaluation of a subpath from node $n$ to node $o$.
We assume that $h(n)$ is FSD-dominated by the random cost of any subpath from node $n$ to node $d$.
Heuristic $h(n)$ can be a deterministic value \cite{ChenLamSumaleeLiTam14} as usual or more generally a random variable \cite{ParmentierMeunier14}.
}




\paul{
Defining the label as such and comparing them with FSD dominance are justified because of the following lemma \cite{NieWu09,ChenLamSumaleeLiTam14,ParmentierMeunier14} and corollary, written for $X, Y, Z$ three real random variables. 
\begin{lemma}\label{lem:add}
If $F_{X} \succsim_{FSD} F_{Y}$, then $F_{X + Z} \succsim_{FSD} F_{Y + Z}$.
\end{lemma}
This lemma can be interpreted in our context as follows: 
If the label (\ie random variable or its associated probability distribution more exactly) of a subpath $\pi$ ending in $n$ FSD-dominates the label of another subpath also ending in $n$, then any extension of those two subpaths will keep the direction of the dominance.
}

\paul{As a direct consequence of Lemma~\ref{lem:add} and FSD, we have:
\begin{corollary}\label{cor:fsd}
If $F_{X} \succsim_{FSD} F_{Y}$, then $\rho(X + Z) \ge \rho(Y + Z)$.
\end{corollary}
This corollary states that if a given node $n$ has two labels, one FSD-dominating the other, the former label can be pruned as it will lead to a higher $\rho$ value.
}

\paul{
Thanks to INC, the following proposition explains why it is sound to end the algorithm as soon as node $d$ is examined (Line~\ref{algol:end} of Algorithm~\ref{algo:aastar}). 
\begin{proposition}
When node $d$ is chosen, the corresponding path is $\rho$-minimum.
\end{proposition}
Property INC was not considered in Parmentier et al.'s work \cite{ParmentierMeunier14}.
Contrary to their algorithm, ours can stop as soon as a path to node $d$ is found.
}

In order to avoid generating too many subpaths, we use an upperbound $UB$ on the best $\rho$ value known so far.
When starting Algorithm~\ref{algo:aastar}, we can use $UB = +\infty$ or better compute a standard shortest path and use its $\rho$ value as an upperbound.
Then, $UB$ can be updated each time a path to $d$ is found (Line~\ref{algol:improveUB}).
Besides, this algorithm can be sped up by pruning with any other known lower bound to the $\rho$ value (see the case study where we use the expected duration).

\paul{
Note that in general, Line~\ref{algol:end} may be hard to compute.
In our case study, we assume the time is discretized into equal-length intervals on which probability distributions are assumed to be constant.
Moreover, we also assume all distributions are discretized.
In the next section, we explain this in more details.
}

\section{Case Study}\label{sec:casestudy}

\paul{We demonstrate our algorithm with $\rho$ chosen as the conditional value-at-risk (CVaR) with $\alpha=90\%$.
This seems to be a better choice than VaR, which was used in Chen et al.'s work \cite{ChenLamSumaleeLiTam14}, because it not only takes into account the VaR threshold, but also the tail distribution.
Besides, being a coherent risk measure \cite{ArtznerDelbaenEberHeath99}, it enjoys nicer properties than VaR.
We implemented our adapted A* algorithm in OpenTripPlanner\footnote{\url{http://www.opentripplanner.org}}, an open-source platform for route planning, which offers a map-based web interface and standard shortest path algorithms.

In order to work with real traffic data, we estimated the dynamic random costs from taxi trip data\footnote{\url{http://www.nyc.gov/html/tlc/html/about/trip_record_data.shtml}} released by the New York City TLC (Taxi and Limousine Commission).
We first explain how a probability distribution for the random duration of an edge was estimated and then present an illustration of results that can be obtained thanks to our algorithm.
}

\subsubsection{Data Cleaning and Estimation.}

\paul{The dataset contains records of taxi trips in Manhattan from 2009 to 2015. 
We only used the data from 2011 to 2015, as the data size was large and we preferred focusing on the most recent records. 
The dataset contains trip information including pick-up/drop-off locations and pick-up/drop-off times.} 
\paul{
We only took into account trips inside the Manhattan area, which represents a network of 5,111 nodes and 16,396 edges.
During the data cleaning phase, we filtered out trips that had a pick-up or drop-off location outside Manhattan. 
We also removed abnormal trips, which may be due to incorrect GPS readings.}

Because the actual path of a trip and detailed times at each intersection of a trip were not provided, we had to make two assumptions to extract random duration $C_t(e)$ of an edge $e \in E$ from the dataset: \\[.7ex]
{\bf ~~~~A1} A trip follows the shortest path from origin to destination. \\[.5ex]
{\bf ~~~~A2} The driver maintains the same speed along the trip. 

\dajian{
\paul{
Given the nature of the dataset, the assumptions seem reasonable enough.
A1 leads to a small overestimation of travel durations in each edge.
For our risk-averse route planning problem, overestimation is better than underestimation.
A2 is a simplifying assumption, which neglects the effects of traffic lights, intersections, turns...
We do not think it has a too big impact for our application, especially given that we have already overestimated the durations.}

Based on A1, for each trip, we computed its shortest path from its origin to its destination using standard A* in terms of duration, where the duration of an edge (\ie portion of a street) equals to the length (\ie distance) of an edge divided by the maximum speed limit allowed in that edge.
Then, given the computed shortest path $\pi$, we could generate a duration sample for each of its edge based on A2 with 
}
$c_e = c_\pi \times \frac{l_e}{l_\pi}$ 
where $c_e$ is the duration of an edge $e$ of $\pi$, $c_\pi$ the total duration of the trip, $l_e$ the length of edge $e$ and $l_\pi$ the length of $\pi$.

Samples $c_e$'s were then collected and used to estimate $P_{C_t(e)}$. 
\paul{As we expect different traffic patterns on weekdays and during weekends}, we divided the days of a week into two classes: 
$Weekdays = \{Mon., Tues., Wed., Thur., Fri.\}$ and 
$Weekends = \{Sat., Sun.\}.$
We divided a day into 24 bins of 1 hour.
For a specific edge $e$, we obtained 24 distributions $P_{C_t(e)}$ (one for each hour) and assume the distribution was constant during an interval of one hour.
Moreover, we assume those distributions are discrete and defined over 100 bins of 6 seconds.
\dajian{
Durations that exceeds 600 seconds were counted as 600 seconds.}

\subsubsection{Experimental Results.}


\dajian{With the adapted A* algorithm described before, we can find the risk-averse path between any pair of origin and destination. 
\paul{For this case study, following Chen et al.'s work \cite{ChenLamSumaleeLiTam14}, we define the heuristic function used in the implemented risk-averse path finding system as 
$h(n) = \frac{d(n)}{v_{max}}$
where $d(n)$ is the shortest length of a path from node $n$ to node $d$ and $v_{max}$ is the maximum travel speed in the network.
}
}
Therefore, $h(n)$ is the shortest possible duration to go from $n$ to $d$.

Besides, it is known that $CVaR_\alpha$ is increasing with $\alpha$ and $CVaR_{0\%}$ is the expectation. 
Therefore, we also maintained an expected duration of a subpath $\pi_{on}$ and estimated a lowerbound of the expected duration of an extension of $\pi_{on}$ to node $d$ (as in standard A*).
This lower bound can be used to prune subpaths by comparing it with the upperbound $UB$.

\begin{figure}[bt]
\centering
\subfigure[path at 6:00 a.m. on Wednesday]{
\centering \includegraphics[width=0.35\linewidth]{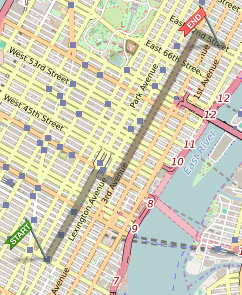}
}
\subfigure[path at 8:00 a.m. on Wednesday]{
\centering \includegraphics[width=0.35\linewidth]{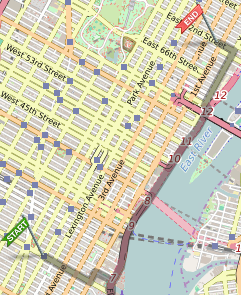}
}
\caption{Examples of risk-averse paths}\label{fig:snapshot}
\end{figure}


\paul{
To illustrate our system, we present one example where for the same pair of origin and destination nodes, CVaR yields different risk-averse paths depending on the departure time. 
As depicted in Figure \ref{fig:snapshot}, at 6:00 a.m. on Wednesday, before rush hours, the risk-averse path (with a CVaR of 24 mins) is very similar to the shortest distance path because there is little risk of congestion. 
Its CVaR can be interpreted as follows: In the worst $10\%$ of the case, the average duration of the trip will be 24 minutes. 
And, in most cases, the observed travel duration would be much less than 24 minutes.
}
In contrast, at 8:00 a.m. on the same day during rush hour, the risk-averse path is no longer the shortest distance path, but a path (with a CVaR of 31 mins) that passes via a highway, which has less probability of congestion. 
Although the risk-averse path may be a longer path to drive, it is a less risky path in terms of CVaR.

The computation times depend on the origin and destination nodes. 
By averaging over 100 runs where those pairs where selected randomly, the average computation time was less than one second (976.2 millisecs) using a computer equipped with an {\em Intel Xeon E31225 @ 3.10GHz}.
To make this system usable in a real application, the computation time could be further improved.
We expect this could be achieved with different optimization techniques: \eg memoization, better heuristics, fitting duration samples to continuous distributions...
As we wanted to demonstrate the feasibility of our approach, we leave this as future work.

\section{Conclusion}\label{sec:conclusion}

\paul{
In this paper, we proposed an adapted A* algorithm, which accommodates any risk measure or decision criterion that is monotonic with first-order stochastic dominance, to find a risk-averse shortest path in a transportation network with time-dependent stochastic costs. 
Besides, we demonstrated our algorithm on a case study with NYC taxi data and obtained reasonable results.


As future work, we plan to improve the computational efficiency of our method, taking inspiration from the techniques developed for standard shortest path problems \cite{BastDellingGoldbergMuller-HannemannPajorSandersWagnerWerneck15,DellingSandersSchultesWagner09}.
Moreover, we would like to test our system on more accurate historical traffic data.
Finally, we plan to extend the approach to take into account other kinds of costs, such as power consumption.


}

\bibliography{biblio160226}

\begin{thebibliography}{10}
\providecommand{\url}[1]{\texttt{#1}}
\providecommand{\urlprefix}{URL }

\bibitem{ArtznerDelbaenEberHeath99}
Artzner, P., Delbaen, F., Eber, J., Heath, D.: Coherent measures of risk.
  Mathematical Finance  9(3),  203--228 (1999)

\bibitem{BastDellingGoldbergMuller-HannemannPajorSandersWagnerWerneck15}
Bast, H., Delling, D., Goldberg, A., M{\"u}ller-Hannemann, M., Pajor, T.,
  Sanders, P., Wagner, D., Werneck, R.: Route planning in transportation
  networks (2015), arXiv:1504.05140v1

\bibitem{BauerleMuller05}
B{\"a}uerle, N., M{\"u}ller, A.: Stochastic orders and risk measures:
  Consistency and bounds. Mathematics \& Economics  (2005)

\bibitem{Bellman58}
Bellman, R.: On a routing problem. Quarterly of Applied Mathematics  16,
  87--90 (1958)

\bibitem{BertsekasTsitsiklis91}
Bertsekas, D., Tsitsiklis, J.: An analysis of stochastic shortest paths
  problems. Mathematics of Operations Research  16,  580--595 (1991)

\bibitem{ChenLamSumaleeLiTam14}
Chen, B.Y., Lam, W.H.K., Sumalee, A., Li, Q., Tam, M.L.: Reliable shortest path
  problems in stochastic time-dependent networks. Journal of Intelligent
  Transportation Systems  18(2),  177--189 (2014)

\bibitem{DellingSandersSchultesWagner09}
Delling, D., Sanders, P., Schultes, D., Wagner, D.: Engineering route planning
  algorithms. Algorithmics, LNCS  5515,  117--139 (2009)

\bibitem{Dijkstra59}
Dijkstra, E.: A note on two problems in connexion with graphs. Numerische
  Mathematik  1,  269--271 (1959)

\bibitem{Dreyfus69}
Dreyfus, S.: An appraisal of some shortest-path algorithms. Operations Research
   17(3),  395--412 (1969)

\bibitem{EmbrechtsKluppelbergMikosch97}
Embrechts, P., Kluppelberg, C., Mikosch, T.: Modelling Extremal Events for
  Insurance and Finance. Springer (1997)

\bibitem{Ford56}
Ford, L.J.: Network flow theory. Tech. rep., Rand Corporation (1956)

\bibitem{Frank69}
Frank, H.: Shortest paths in probabilistic graphs. Operations Research  17(4),
  583--599 (1969)

\bibitem{FuRilett98}
Fu, L., Rilett, L.: Expected shortest paths in dynamic and stochastic traffic
  networks. Transportation Research Part B: Methodological  32(7),  499--516
  (1998)

\bibitem{GavrielHanasusantoKuhn12}
Gavriel, C., Hanasusanto, G., Kuhn, D.: Risk-averse shortest path problems. In:
  IEEE 51st Annual Conference on Decision and Control. pp. 2533--2538 (2012)

\bibitem{GoldbergHarrelson05}
Goldberg, A., Harrelson, C.: Computing the shortest path: A* meets graph
  theory. In: SODA. pp. 156--165 (2005)

\bibitem{HartNilssonRaphael68}
Hart, P.E., Nilsson, N.J., Raphael, B.: A formal basis for the heuristic
  determination of minimum cost paths. IEEE Trans. Syst. and Cyb.  4(2),
  100--107 (1968)

\bibitem{Jorion06}
Jorion, P.: Value-at-Risk: The New Benchmark for Managing Financial Risk.
  McGraw-Hill (2006)

\bibitem{KaufmanSmith93}
Kaufman, D., Smith, R.: Fastest paths in time-dependent networks for
  intelligent vehicle-highway systems application. Journal of Intelligent
  Transportation Systems  1(1),  1--11 (1993)

\bibitem{Moore59}
Moore, E.F.: The shortest path through a maze. In: Proc. of the Int. Symp. on
  the Theory of Switching. pp. 285--292 (1959)

\bibitem{vonNeumannMorgenstern44}
von Neumann, J., Morgenstern, O.: Theory of games and economic behavior.
  Princeton university press (1944)

\bibitem{NieWu09}
Nie, Y., Wu, X.: Shortest path problem considering on-time arrival probability.
  Transportation Research Part B: Methodological  43(6),  597--613 (2009)

\bibitem{OgryczakRuszczynski99}
Ogryczak, W., Ruszczynski, A.: From stochastic dominance to mean-risk models:
  Semideviations as risk measures. European Journal of Operational Research
  116(33--50) (1999)

\bibitem{OrdaRom90}
Orda, A., Rom, R.: Shortest-path and minimum delay algorithms in networks with
  time-dependent edge-length. Journal of the ACM  37(3),  607--625 (1990)

\bibitem{ParmentierMeunier14}
Parmentier, A., Meunier, F.: Stochastic shortest paths and risk measures. In:
  arXiv preprint (2014)

\bibitem{PeyerRautenBachVygen09}
Peyer, S., RautenBach, D., Vygen, J.: A generalization of dijkstra's shortest
  path algorithm with applications to vlsi routing. Journal of Discrete
  Algorithms  7(4),  377--390 (2009)

\bibitem{Puterman94}
Puterman, M.: Markov decision processes: discrete stochastic dynamic
  programming. Wiley (1994)

\bibitem{Quiggin93}
Quiggin, J.: Generalized expected utility theory: the rank-dependent model.
  Kluwer Academic Publishers (1993)

\bibitem{RussellNorvig03}
Russell, S., Norvig, P.: Artificial Intelligence: A Modern Approach.
  Prentice-Hall, 2nd edn. (2003)

\bibitem{Savage54}
Savage, L.: The foundations of statistics. J. Wiley and sons (1954)

\bibitem{ShakedShanthikumar94}
Shaked, M., Shanthikumar, J.: Stochastic Orders and Their Applications.
  Academic press (1994)

\bibitem{SigalPritskerSolberg80}
Sigal, C., Pritsker, A., Solberg, J.: The stochastic shortest route problem.
  Operations Research  28,  1122--1129 (1980)

\bibitem{Yaari87}
Yaari, M.: The dual theory of choice under risk. Econometrica  55,  95--115
  (1987)

\end{thebibliography}
\bibliographystyle{splncs03}

\end{document}